\pdfoutput=1

\documentclass[11pt]{article}

\usepackage{EMNLP2022}

\usepackage{times}
\usepackage{latexsym}

\usepackage[T1]{fontenc}

\usepackage[utf8]{inputenc}

\usepackage{inconsolata}
\usepackage{microtype}
\usepackage{amsmath}
\usepackage{amssymb}
\usepackage{booktabs}
\usepackage{multirow}
\usepackage{graphicx}  
\usepackage{subfig}    

\title{SMiLE: Schema-augmented Multi-level Contrastive Learning for Knowledge Graph Link Prediction}

\newcommand*{\affaddr}[1]{#1} 
\newcommand*{\affmark}[1][*]{\textsuperscript{#1}}
\newcommand*{\email}[1]{\texttt{#1}}

\author{
Miao Peng\affmark[1], Ben Liu\affmark[1], Qianqian Xie\affmark[2], Wenjie Xu\affmark[1], Hua Wang\affmark[3], Min Peng\affmark[1]\thanks{*Corresponding author}\\
\affaddr{\affmark[1]School of Computer Science, Wuhan University, China}\\
\affaddr{\affmark[2]Department of Computer Science, The University of Manchester, United Kingdom}\\
\affaddr{\affmark[3]Centre for Applied Informatics, Victoria University, Australia}\\
\email{\{pengmiao,liuben123,vingerxu,pengm\}@whu.edu.cn}\\
\email{qianqian.xie@manchester.ac.uk,hua.wang@vu.edu.au}
}

\begin{document}
\maketitle
\begin{abstract}

Link prediction is the task of inferring missing links between entities in knowledge graphs. Embedding-based methods have shown effectiveness in addressing this problem by modeling relational patterns in triples. However, the link prediction task often requires contextual information in entity neighborhoods, while most existing embedding-based methods fail to capture it. Additionally, little attention is paid to the diversity of entity representations in different contexts, which often leads to false prediction results. In this situation, we consider that the schema of knowledge graph contains the specific contextual information, and it is beneficial for preserving the consistency of entities across contexts. In this paper, we propose a novel \textbf{S}chema-augmented \textbf{M}ult\textbf{i}-level contrastive \textbf{LE}arning framework (SMiLE) to conduct knowledge graph link prediction. Specifically, we first exploit network schema as the prior constraint to sample negatives and pre-train our model by employing a multi-level contrastive learning method to yield both prior schema and contextual information. Then we fine-tune our model under the supervision of individual triples to learn subtler representations for link prediction. Extensive experimental results on four knowledge graph datasets with thorough analysis of each component demonstrate the effectiveness of our proposed framework against state-of-the-art baselines. The implementation of SMiLE is available at \href{https://github.com/GKNL/SMiLE}{https://github.com/GKNL/SMiLE}.

\end{abstract}

\section{Introduction}
Knowledge graph (KG), as a well-structured representation of knowledge, stores a vast number of human knowledge in the format of triples-(head, relation, tail). 
KGs are essential components for various artificial intelligence applications, including question answering~\citep{Diefenbach_QA}, recommendation systems~\citep{KGIN}, etc. In real world, KGs always suffer from the incompleteness problem, meaning that there are a large number of valid links in KG are missing. In this situation, link prediction techniques, which aim to automatically predict whether a relationship exists between a head entity and a tail entity, are essential for triple construction and verification.

\begin{figure}[!t]
    \centering
    \includegraphics[width=3in]{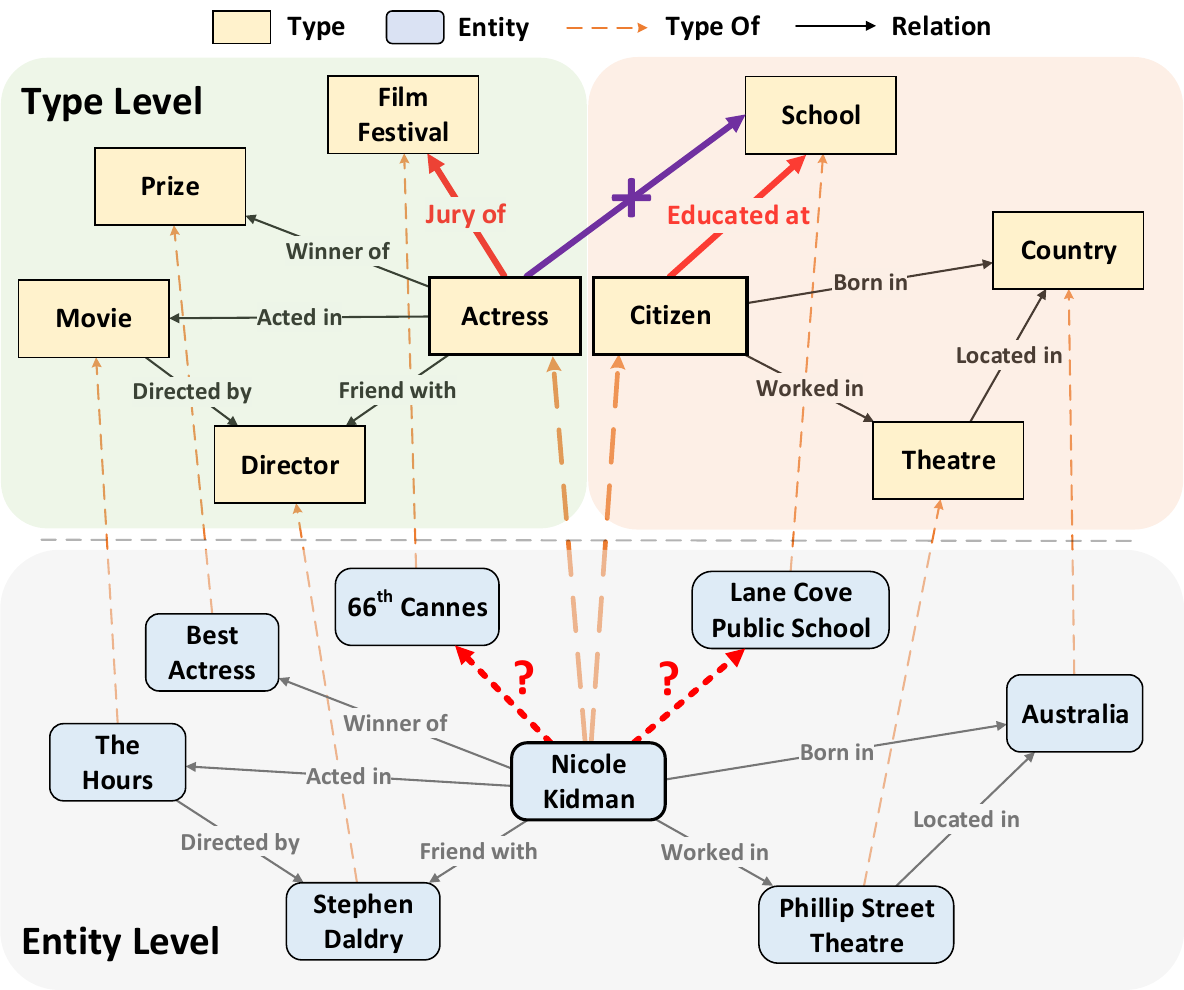}
    \caption{An example of KG fragment. \emph{Nicole Kidman} has two types \emph{Actress} and \emph{Citizen}, and each of them preserves different information in different contexts.}
    \label{fig:Schema Example}
\end{figure}

To address the link prediction problem in KG, a variety of methods have been proposed. Traditional rule-based methods like Markov logic networks~\citep{Markov_logic} and reinforcement learning-based method~\citep{DeepPath} learn logic rules from KGs to conduct link prediction. The other mainstream methods are based on knowledge graph embeddings, including translational models like TransE~\citep{TransE}, TransR~\citep{TransR} and semantic matching models like RESCAL~\citep{RESCAL}, DistMult~\citep{DistMult}. Besides, embedding-based methods leverage graph neural networks to explore graph topology~\citep{CompGCN} and utilize type information~\citep{TransT} to enhance representations in KG.

Nevertheless, the aforementioned methods fail to model the contextual information in entity neighbors. In fact, the context of an entity preserves specific structural and semantic information, and link prediction task is essentially dependent on the contexts related to specific entities and triples.
Furthermore, not much attention is paid to the diversity of entity representations in different contexts, which may often result in false predictions. Quantitatively, dataset FB15k has 14579 entities and 154916 triples, and the number of entities with types is 14417 (98.89\%). There are 13853 entities (95.02\%) that have more than two types, and each entity has 10.02 types on average. For example, entity \textit{Nicole Kidman} in Figure~\ref{fig:Schema Example} has two different types (\textit{Actress} and \textit{Citizen}), expressing different semantics in two different contexts. Specifically, the upper left in the figure describes the contextual information in type level about "Awards and works of \textit{Nicole Kidman} as an \textit{actress}". In this case, it is well-founded that there exists a relation between \textit{Nicole Kidman} and \textit{66$^{th}$ Cannes}, and intuitively the prediction of \textit{(Nicole Kidman, ?, Lane Cove Public School)} does not make sense, since there is no direct relationship between type \textit{Actress} and type \textit{School}. But considering that \textit{Nicole Kidman} is also an Australian \textit{citizen}, it is hence reasonable to conduct such a prediction.

We argue that the key challenge of preserving contextual information in embeddings is how to encapsulate complex contexts of entity neighborhoods. Simply considering all information in the subgraph of entities as the context may bring in redundant and noisy information. Schema, as a high-order meta pattern of KG, contains the type constraint between entities and relations, and it can naturally be used to capture the structural and semantic information in context. As for the problem of inconsistent entity representations, the diverse representations of an entity are indispensable to be considered in different contexts. As different schema defines diverse type restrictions between entities, it is able to preserve subtle and precise semantic information in a specific context.
Additionally, to yield consistent and robust entity representations for each contextual semantics, entities in contexts of the same schema are supposed to contain similar features but disparate in different contexts.

To tackle the aforementioned issues, inspired by the advanced contrastive learning techniques, we proposed a novel schema-augmented multi-level contrastive learning framework to allow efficient link prediction in KGs.
To tackle the incompleteness problem of KG schema, we first extract and build a \textit{<head\_type, relation, tail\_type>} tensor from an input KG~\citep{RETA} to represent the high-order schema information.
Then, we design a multi-level contrastive learning method under the guidance of schema. Specifically, we optimize the contrastive learning objective in contextual-level and global-level of our model separately.
In the contextual-level, contrasting entities within subgraphs of the same schema can learn semantic and structural characteristics in a specific context. In the global-level, differences and global connections between contexts of an entity can be captured via a cross-view contrast.
Overall, we exploit the aforementioned contrastive strategy to obtain entity representations with structural and high-order semantic information in the pre-train phase and then fine-tune representations of entities and relations to learn subtler knowledge of KG.

To summarize, we make three major contributions in this work as follows:
\begin{itemize}
     \item We propose a novel multi-level contrastive learning framework to preserve contextual information in entity embeddings. Furthermore, we learn different entity representations from different contexts.
    \item We design a novel approach to sample hard negatives by utilizing KG schema as a prior constraint, and perform the contrast estimation in both contextual-level and global-level, enforcing the embeddings of entities in the same context closer while pushing apart entities in dissimilar contexts.
    \item We conduct extensive experiments on four different kinds of knowledge graph datasets and demonstrate that our model outperforms state-of-the-art baselines on the link prediction task.
\end{itemize}

\section{Related Work}
\subsection{KG Inference}
To conduct inference like link prediction on incomplete KG, most traditional methods enumerate relational paths as candidate logic rules, including Markov logic network~\citep{Markov_logic}, rule mining algorithm~\citep{BURL_KGC} and path ranking algorithm~\citep{RWI_KB}. However, these rule-based methods suffer from limited generalization performance due to consuming searching space.

The other mainstream methods are based on reinforcement learning, which defines the problem as a sequential decision-making process~(\citealp{DeepPath}; \citealp{Multi-Hop_Reasoning_RL}). They train a pathfinding agent and then extract logic rules from reasoning paths. However, the reward signal in these methods can be exceedingly sparse.

\subsection{KG Embedding Models}
Various methods have been explored yet to perform KG inference based on KG embeddings. Translation-based models including TransE~\citep{TransE}, TransR~\citep{TransR} and RotatE~\citep{RotatE} model the relation as a translation operation from head entity to tail entity. Semantic matching methods like DistMult~\citep{DistMult} and QuatE~\citep{QuatE} measure the authenticity of triples through a similarity score function. GNN-based methods are proposed to comprehensively exploit structural information of neighbors by a message-passing mechanism. R-GCN~\citep{R-GCN} and CompGCN~\citep{CompGCN} employ GCNs to model multi-relational KG. 

More recently, some methods integrate auxiliary information into KG embeddings. JOIE~\citep{JOIE} considers ontological concepts as supplemental knowledge in representation learning. TransT~\citep{TransT} and TKRL~\citep{TKRL} leverage rich information in entity types to enhance representations. Nevertheless, these graph-based methods further capture relational and structural information but fail to capture the contextual semantics and schema information in KG.

\subsection{Graph Contrastive Learning}
Contrastive learning is an effective technique to learn representation by contrasting similarities between positive and negative samples \citep{CRL2020}.
More recently, the self-supervised contrastive learning method has been introduced into graph representation area. HeCo~\citep{HeCo} proposes a co-contrastive learning strategy for learning node representations from the meta-path view and schema view. CPT-KG~\citep{CPT-KG} and PTHGNN~\citep{PTHGNN} optimize contrastive estimation on node feature level to pre-train GNNs on heterogeneous graphs. Furthermore, \citet{CCC} proposes a  hierarchical contrastive model to deal with representation learning on imperfect KG. SimKGC~\citep{SimKGC} explores a more effective contrastive learning method for text-based knowledge representation learning with pre-trained language models.

\section{The Proposed SMiLE Framework}

\begin{figure*}[!t]
    \centering
    \includegraphics[width=6.25in]{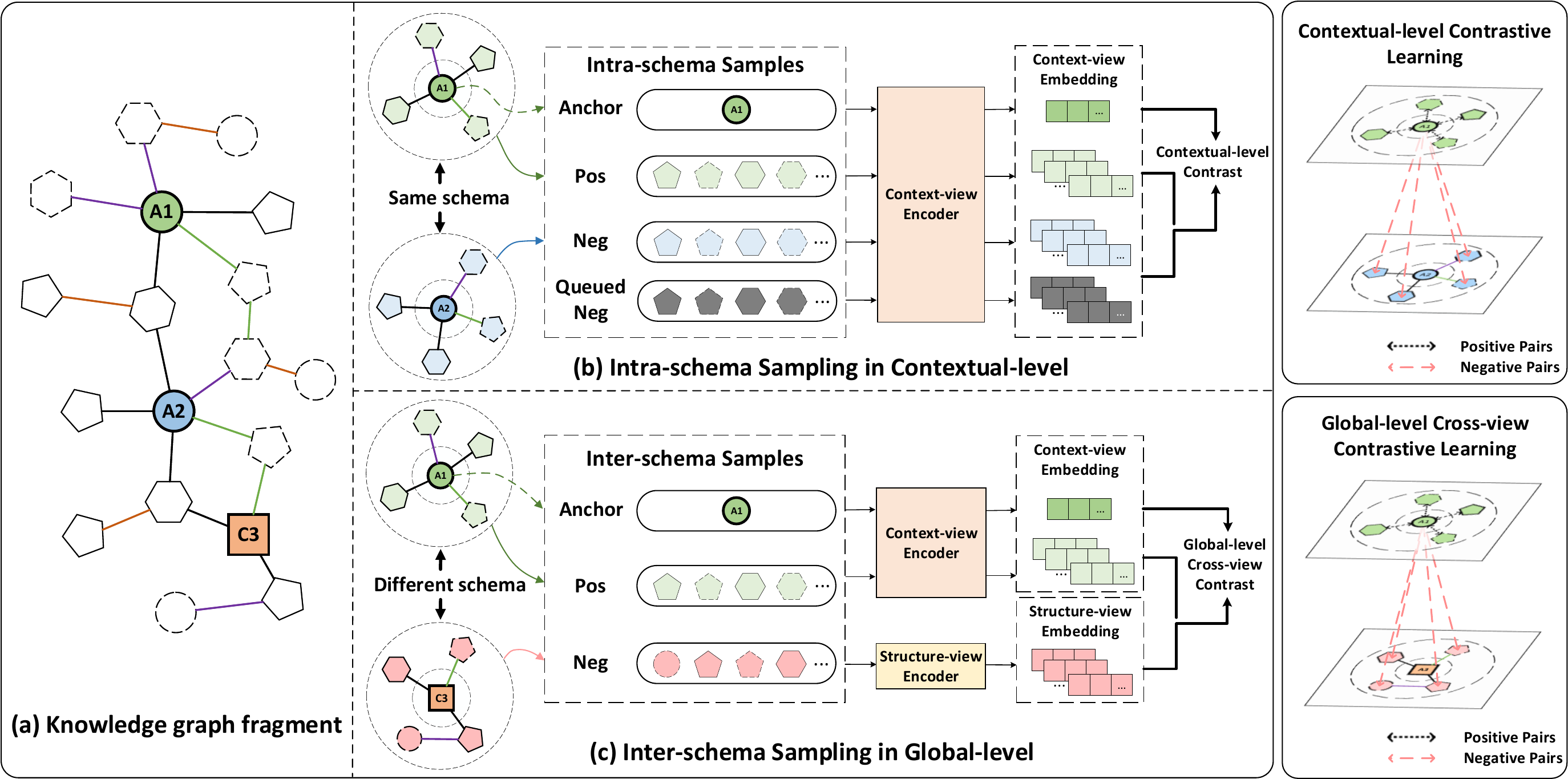}
    \caption{Overall illustration of the proposed SMiLE model: detailed framework of SMiLE model(left) and a sketch map of multi-level contrastive learning mechanism(right).}
    \label{fig:Overall Framework}
\end{figure*}

In this section, we first present notations related to this work. Then we introduce the detail and training strategy of our proposed framework. The overall architecture of SMiLE is shown in Figure~\ref{fig:Overall Framework}.

\subsection{Notations}
A knowledge graph can be defined as $\cal{G = (E, R, T, P)}$, where $\cal{E}$ and $\cal{R}$ indicate the set of entities and relations, respectively. $\cal{T}$ represents the collection of triples $(s, r, o)$ and $\cal{P}$ is the set of all entity types. Each entity $s (or\ o) \in \cal{E}$ has one or multiple types $t_{s1}, t_{s2}, ..., t_{sn} \in \cal{P}$.

The goal of our SMiLE model is to study the structure- and context-preserving properties of entity representations to perform effective link prediction tasks in knowledge graphs, which aim to infer missing links in an incomplete $\cal G$. Ideally, the probability scores of positive triples are supposed to be higher than those of corrupted negative ones.

\noindent \textbf{Context Subgraph.} Given an entity $s$, we regard its $k$-hop neighbors with related edges as its context subgraph, denoted as $g_c(s)$. Likewise, we define the context subgraph between two entities $s$ and $o$ as the $k$-hop neighbors connecting s and o via several relations, which can be represented as $g_c(s,o)$.

\noindent \textbf{Knowledge Graph Schema.} The schema of KG can be defined as $S = (\cal{P},\cal{R})$, where $\cal{P}$ is the set of all entity types and $\cal{R}$ is the set of all relations. Consequently, the schema of a KG can be characterized as a set of entity-typed triples $(t_s, r, t_o)$, meaning that entity $s$ of type $t_s$ has a connection with entity $o$ of type $t_o$ via a relation $r$. 

\subsection{Network Schema Construction} \label{Schema Construction}
By reason of some existing KGs do not contain complete schema, inspired by RETA~\citep{RETA}, we design a simple but effective approach to construct schema $\cal{S}$ from a KG $\cal{G}$.

First, for all triples $(s,r,o)$ in KG, we convert each entity to its corresponding type, hence all entity-typed triples form a typed collection $S = \{(t_s,r,t_o)|(t_s,r,t_o)\in \cal{P} \times \cal{R} \times \cal{P}\}$. Noticing that each entity in KG may have multiple types, we take each combination of entity types in an entity-typed triple into consideration. Then, we calculate the frequency of each entity-typed triple and filter out those with a frequency below threshold $\alpha$, which indicates few contributions to the schema in KG. Finally, we obtain the KG schema represented in the form of a boolean tensor $T \in \mathbb{B}^{|P^s| \times |R| \times |P^o|}$, where $P^s$ and $P^o$ are the set of filtered head and tail types respectively.

\noindent \textbf{Context Schema.} Given an entity $s$ and its context subgraph $g_c(s)$, we get an entity-typed subgraph $S_t(s)$ by converting all entities in $g_c(s)$ to their corresponding types. Then we apply the intersection operation between $S_t(s)$ and KG schema $S$, hence we obtain the context schema of $g_c(s)$ as:
\[
    S_c(s) = \big \{ (t_s,r,t_o)\ |\ (t_s,r,t_o) \in S_t(s) \cap S \big \}.
\]

\subsection{Multi-view Entity Encoder} \label{Multi-view Encoder}
Generally, entities preserve multiple expressions under different views, hence we encode entities into different representations to preserve diverse features in context- and structure-view, respectively.

\noindent \textbf{Structure-view Encoder.} Given an entity $s$ and a relation $r$, we first obtain their global structure-aware representations as follows:

\begin{equation*}
h_s = f_e(s;{\cal G}); z_r = f_r(r).
\end{equation*}

To obtain graph-structure based embeddings in KG, we adopt the GNN model as the implementation of $f_e(\cdot ;\cal{G})$ and we use the $i.i.d.$ embedding network to implement $f_r(\cdot)$.

\noindent \textbf{Context-view Encoder.}
To capture inherent knowledge in a context schema, we employ the $k$-layer stacked contextual translation function~\citep{SLiCE} to learn contextual embeddings of entities ${\cal E}_c$ with embeddings $H_c=(h_1,h_2, ..., h_{|\mathcal{E}_c|}) \in \mathbb{R}^{d \times |{\cal E}_c|}$ in a context subgraph $g_c$ of entity $s$:
\begin{equation*}
H_c^{i+1} = \textrm{Enc}( W_{sc} H_c^i \bar{A}^i + H_c^i), i=0,1,...,k-1,
\end{equation*}
where $\textrm{Enc(·)}$ is a MLP encoder, $W_{sc} \in \mathbb{R}^{d_k \times d_{k+1}}$ is a layer-specific parameter matrix and $\bar{A}^i \in \mathbb{R}^{|{\cal E}_c| \times |{\cal E}_c|}$ denotes the semantic association matrix which is computed by multi-head attention mechanism. Then we get the context-view embedding of node $s$ by aggregating the output of each layer:
\begin{equation*}
c_s = h_s^0 \oplus h_s^1 \oplus ... \oplus h_s^{k-1}.
\end{equation*}

\subsection{Contextual-level Contrastive Learning} \label{Contextual-level CL}
We exploit contextual-level contrastive learning to capture latent semantics and correlations of entities within context schemas. A context schema constrains the type of tail entities and relations that a head entity can be related to, and it is helpful to obtain harder negative samples, contributing to more effective contrast estimation.

\noindent \textbf{Positive Samples.}
Given a context subgraph $g_c(s)$ and its corresponding context schema $S_c(s)$, let $s$ be the anchor entity of $g_c(s)$ and $S_c(s)$, while the others in $g_c(s)$ be the context entities. We define the positive samples of anchor entity $s$ in both contextual-level and global-level as follows:
\[
{\cal P}_s = \big \{u\ |\ (s,r,u) \in {\cal T}_c(s), u \neq s \big \},
\]
where ${\cal T}_c(s)$ is the set of triples in $g_c(s)$.

\noindent \textbf{Intra-schema Negative Samples.} For two anchor entities $u$ and $v$ matching the same type, if context subgraphs $g_c(u)$ and $g_c(v)$ generated from them can be projected to the same context schema, we define their neighbor entities within $g_c(s)$ as negative samples of each other. Formally, given a batch of anchor entities ${\cal E}_B$, we denote the negative samples of entity $s$ as:

\begin{small}
\[
{\cal N}_s^{cur} = \big \{{\cal P}_i\ |\ i \in {\cal E}_B \setminus \{s\}, S_c(i) = S_c(s), t_i = t_s \big \}.
\]
\end{small}

Generally, the number of intra-schema negative samples in a batch is usually coupled with batch size. We employ a dynamic queue to store entity embeddings from previous batches~(\citealp{MoCo}; \citealp{SimKGC}), aiming to increase the number of intra-schema negative samples. We denote the queued pre-batch negative samples of entity $s$ as:

\begin{equation*}
\begin{split}
{\cal N}_s^{pre} = \big \{{\cal P}_j\ |\ j \in {\cal E}_B^{-1} \cup {\cal E}_B^{-2} ... \cup {\cal E}_B^{-n}, \\
S_c(j) = S_c(s), t_j = t_s \big \},
\end{split}
\end{equation*}
where ${\cal E}_B^{-n}$ represents entities of $n$-th pre-batch. Since embeddings from previous batches are computed with previous model parameters, we usually limit $n$ with a small number to ensure that they are consistent with negative samples in ${\cal N}_s^{cur}$. The total intra-schema negative samples of anchor entity $s$ in contextual-level are:
\[
{\cal N}_s^{ia} = {\cal N}_s^{cur} \cup {\cal N}_s^{pre}.
\]

\noindent \textbf{Contextual-level Optimization.} With context-view embeddings of entities, we apply InfoNCE loss~\citep{CCC} to perform contrast estimation as follows:

\begin{equation*}
\begin{split}
\mathcal{L}^{c}_s = -\log \frac{\sum\limits_{t \in \mathbb{P}_{s}} \exp \left (\phi(c_s, c_t) / \tau\right)}{\sum\limits_{k \in \{\mathbb{P}_{s} \cup \mathbb{N}_{s}^{ia}\}} \exp \left(\phi(c_s, c_k) / \tau\right)},
\end{split}
\end{equation*}

where $\tau$ is the temperature hyper-parameter to control the sensitivity of score function, and we apply cosine similarity as the score function $\phi$. Different from previous contrast-based methods, we take multiple positive samples into consideration in computing contrastive loss.

\subsection{Global-level Contrastive Learning} \label{Global-level CL}
In addition to local contexts, it is essential to capture correlations among various context subgraphs. We apply the cross-view contrastive learning strategy to strike a balance between global schema and contextual features of KG representations.

\noindent \textbf{Inter-schema Negative Samples.} If $u$ and $v$ are two anchor entities corresponding to two different context schemas, we define their context entities as negative samples of each other:

\begin{equation*}
\begin{split}
{\cal N}_s^{ie} = \big \{{\cal E}_c^{i}\ |\ i \in {\cal E}_B \setminus \{s\} &,\ S_c(i) \neq S_c(s) \},
\end{split}
\end{equation*}
where ${\cal E}_B$ indicates a batch of anchor entities.

\noindent \textbf{Global-level Optimization.} Obtaining the embeddings of entity $s$ under context- and structure-view, we feed them into an MLP encoder with one hidden layer, hence they are mapped into the space where the contrastive loss is calculated:
\begin{equation*}
\begin{split}
h_s^p &= W^2 \sigma(W^1 h_s + b^1)+b^2, \\
c_s^p &= W^2 \sigma(W^1 c_s + b^1)+b^2,
\end{split}
\end{equation*}
where $\sigma$ is ELU activation function. It is worth noting that weight matrix \{$W^1$, $W^2$\} and bias parameter \{$b^1$, $b^2$\} are shared with embeddings of two different views.
Then we perform cross-view contrastive learning between context- and structure-view representations of entities as follows:

\begin{small} 
\begin{equation*}
\begin{split}
\mathcal{L}_s^{g}= -\log \frac{\sum\limits_{t \in \mathbb{P}_{s}} e^{\left (\phi(c_s^p, c_t^p) / \tau\right)}}{\sum\limits_{k \in \mathbb{P}_{s}} e^{\left(\phi(c_s^p, c_k^p) / \tau\right)}+\sum\limits_{k \in \mathbb{N}_{s}^{ie}} e^{\left(\phi(c_s^p, h_k^p) / \tau\right)}},
\end{split}
\end{equation*}
\end{small}

where $\tau$ is the temperature hyper-parameter and $\phi$ is the cosine similarity score function.

\subsection{Training Objective for link prediction} \label{Training Objective}

To capture both semantic and structural information in the context schema and individual triple for link prediction, we employ a pre-train \& fine-tune pipeline to optimize our proposed SMiLE model.

\subsubsection{Contrastive optimization in pre-training}
In pre-train phase, we employ the multi-level contrastive learning strategy mentioned in \ref{Contextual-level CL} and \ref{Global-level CL} to optimize model parameters $\theta$. To capture semantic and structural knowledge of entities in both contextual- and global-level, we jointly minimize the contextual- and global-level loss as follows:
\[
\mathcal{L} = \frac{1}{|\cal E|} \sum_{s \in \cal E}\left [ \lambda \cdot \mathcal{L}_s^g + (1 - \lambda) \cdot \mathcal{L}_s^c \right ],
\]
where $\lambda$ is a balancing coefficient that controls the weight of two losses under different levels.

\subsubsection{Fine-tuning for link prediction}
With pre-trained model parameters $\theta$ as an initialization, we further fine-tune the model to learn subtler representations of individual entities and relations under the supervision of each individual triple. For a positive triple in KG, we construct its negative samples by corrupting the head or tail entity, with the restriction that the replaced entity should have the same type as the original one. Then, for each triple $(s,r,o)$, we obtain the relation-aware embedding of head entity $s$ as $h_s^r = \Phi (h_s,z_r)$, where $\Phi (\cdot)$ denotes the non-parameterized entity-relation composition operation \citep{CompGCN}, which can be subtraction, multiplication, circular-correlation, etc.

Next, for a triple $(s,r,o)$, we generate its corresponding context subgraph $g_c(s,o)$ by employing a shortest path strategy, which considers the shortest path between entity $s$ and entity $o$ as the context. Feeding entities in $g_c(s,o)$ with their relation-aware embeddings into context-view encoder, we obtain the context-view embeddings of entity $s$ and $o$ in triple $(s,r,o)$, denoted as $c_{s}^r$ and $c_o$.

The training objective in fine-tune phase is as follows:
\begin{equation*}
\begin{split}
{\cal L} &= \sum_{(s,r,o)\in T_p}\log \left(\phi_r(c_{s}^r,c_o)\right) \\
&+ \sum_{(s',r,o')\in T_n}\log \left(1-\phi_r(c_{s'}^r,c_{o'})\right),
\end{split}
\end{equation*}
where $T_p$ and $T_n$ represent the set of positive and negative triples, respectively. $\phi_r(c_s,c_o)$ denotes the score function to measure the compatibility between entities pair via relation r. Here we adopt to the dot product similarity as $\phi_r(c_s,c_o)=\sigma(c_s \cdot c_o^T)$, where $\sigma$ is the sigmoid function.

\subsection{Complexity Analysis}
Theoretically, the major difference between SMiLE and previous baseline models is the negative sampling and contrastive loss, which is related to the number of negative samples. The complexity of the pre-training phase is ${\cal O}(|{\cal E}| \ast |{\cal P}| \ast (k_1 + k_2))$, where $|\cal{E}|$ is the number of entity nodes, $|\cal{P}|$ denotes the size of positive samples in both contextual-level and global-level, $k_1$ and $k_2$ are the number of negative samples per positive sample in contextual-level and global-level. The complexity of fine-tuning phase is ${\cal O}(|R| \ast N_c)$, where $|\cal{R}|$ is the number of relation edges and $N_c$ denotes the maximum number of nodes in any context subgraph.

\section{Experiments}

\subsection{Experimental Setup}

\noindent \textbf{Datasets.} We evaluate our model on four synthetic and real-world KG datasets: FB15k~\citep{TransE}, FB15k-237~\citep{toutanova-chen-2015_FB15k-237}, JF17k~\citep{WenLMCZ16_JF17k}, and HumanWiki~\citep{RETA}. More precisely, FB15k and JF17k are subsets of Freebase, and FB15k-237 is a pruned version of FB15k. HumanWiki is extracted from Wikidata by extracting all triples involving a head entity of type \textit{human}. 
For entity type information, we use the data built in \citet{RETA} and we generate equal number of positive and negative edges for the link prediction task. Statistics of these four datasets are shown in Table~\ref{tab:datasets}.

\begin{table}[!t]\small
\renewcommand{\arraystretch}{1.2}
\centering
\setlength{\tabcolsep}{1.3mm}
\begin{tabular}{l|cccc}
\toprule
\textbf{Dataset} & \textbf{FB15k-237} & \textbf{FB15k} & \textbf{JF17k} & \textbf{HumanWiki}\\
\midrule
\#Entities & 14,541 & 14,579 & 9,233 & 38,949 \\
\#Types & 583 & 588 & 511 & 388 \\
\#Relations & 237 & 1,208 & 326 & 221 \\
\#Edges & 248,611 & 117,580 & 18,049 & 105,688 \\
\#Triples & 310,116 & 154,916 & 19,342 & 108,199 \\
\bottomrule
\end{tabular}
\caption{Statistics of datasets used in this paper.}
\label{tab:datasets}
\end{table}

\noindent \textbf{Baselines.} We compare the proposed SMiLE against nine representative models, which can be divided into two categories. The first category is KGE-based models including TransE~\citep{TransE}, ComplEx-N3~\citep{ComplEx-N3}, TransR~\citep{TransR}, TypeComplex~\citep{TypeComplEx} with additional type information, SANS~\citep{SANS} with structure-aware negative samples and SOTA model PairRE~\citep{pairRE} with paired relation vectors. The second category is GNN-based models that employ a GNN model to exploit structural information in KG, including random-walk based homogeneous network Node2vec~\citep{Node2vec}, multi-relational model CompGCN~\citep{CompGCN}, and SOTA approach SLiCE~\citep{SLiCE} with subgraph-based contextualization.

\begin{table*}[!t]\small
\renewcommand{\arraystretch}{1.1}
 \centering
 \setlength{\tabcolsep}{1.4mm}
 \begin{tabular}{ccccccccc}\toprule
    \multirow{2}{*}{\textbf{Model}} & \multicolumn{2}{c}{\textbf{FB15k}} & \multicolumn{2}{c}{\textbf{FB15k-237}} & \multicolumn{2}{c}{\textbf{JF17k}} & \multicolumn{2}{c}{\textbf{HumanWiki}}
    \\\cmidrule(lr){2-3}\cmidrule(lr){4-5}\cmidrule(lr){6-7}\cmidrule(lr){8-9}
             & F1 & AUC-ROC  & F1 & AUC-ROC  & F1 & AUC-ROC  & F1 & AUC-ROC\\\midrule \specialrule{0em}{1.5pt}{1.5pt}\midrule  
    TransE~\citep{TransE} & 50.36 & 50.13 & 47.78 & 48.18 & 44.68 & 46.18 & 49.06 & 49.31 \\
    TransR~\citep{TransR} & 71.96 & 76.96 & 67.19 & 70.76 & 62.06 & 68.14 & 61.56 & 66.54 \\
    ComplEx-N3~\citep{ComplEx-N3} & 49.63 & 49.63 & 50.19 & 50.34 & 48.44 & 49.15 & 54.53 & 52.86 \\
    TypeComplex~\citep{TypeComplEx} & 88.09 & 93.90 & 50.05 & 50.25 & 73.73 & 78.53 & 80.17 & 85.58 \\
    SANS~\citep{SANS} & \underline{88.97} & 94.59 & 50.03 & 50.26 & 68.62 & 79.41 & 78.18 & 83.69 \\
    PairRE~\citep{pairRE} & 88.27 & 92.67 & 49.62 & 49.30 & 71.89 & 79.65 & 80.07 & 87.68 \\\midrule
    Node2vec~\citep{Node2vec} & 80.23 & 88.91 & 83.69 & 89.77 & 93.30 & 98.01 & 80.13 & 87.54 \\
    CompGCN~\citep{CompGCN} & 60.35 & 63.59 & 65.39 & 72.01 & 66.27 & 52.13 & 56.88 & 40.09 \\
    SLiCE~\citep{SLiCE} & 88.34 & \underline{94.66} & \textbf{90.26} & \textbf{96.41} & \underline{96.16} & \underline{98.89} & \underline{88.92} & \underline{96.19} \\\midrule
    \textbf{SMiLE(ours)} & \textbf{90.76} & \textbf{96.53} & \underline{88.75} & \underline{94.92} & \textbf{96.98} & \textbf{99.22} & \textbf{93.40} & \textbf{97.92}
    \\\bottomrule
 \end{tabular}
 \caption{Link prediction performance of our method(SMiLE) and recent models on FB15k-237, FB15k, JF17k and HumanWiki datasets.
 The best results are in \textbf{bold} and the second best results are \underline{underlined}.}
 \label{tab:comparision results}
\end{table*}

\begin{table*}[!t]\small
 \centering
 \setlength{\tabcolsep}{1.8mm}
 \begin{tabular}{cc|cc|cc|cc}\toprule
    \multicolumn{2}{c|}{\textbf{Model}} & \multicolumn{2}{c|}{\textbf{FB15k}} & \multicolumn{2}{c|}{\textbf{JF17k}} & \multicolumn{2}{c}{\textbf{HumanWiki}}
    \\\midrule
    \textbf{contextual} & \textbf{global}  & F1 & AUC-ROC  & F1 & AUC-ROC  & F1 & AUC-ROC\\\midrule \specialrule{0em}{1.5pt}{1.5pt}\midrule  
    $\checkmark$ & & 90.57 & 96.40 & 96.78 & 99.16 & 92.88 & 97.75 \\
    & $\checkmark$ & \textbf{91.04} & 96.49 & 96.44 & 98.96 & 91.33 & 96.90 \\
    $\checkmark$ & $\checkmark$ & 90.76 & \textbf{96.54} & \textbf{96.98} & \textbf{99.22} & \textbf{93.40} & \textbf{97.92}
    \\\bottomrule
 \end{tabular}
 \caption{Ablation study results on FB15k, JF17k and HumanWiki datasets. The best results are in \textbf{bold}.}
 \label{tab:Ablation Study}
\end{table*}

\noindent \textbf{Implementation Details.} We implement our SMiLE with Pytorch and adopt Adam as the optimizer to train our model with the learning rate of 1e-4 for pre-train phase and 1e-3 for fine-tune phase. Models are trained on NVIDIA TITAN V GPUs. We utilize the random walk approach to generate subgraphs, and the embedding dimension is set to 128. Temperature $\tau$ is initialized to 0.8 and the number of contextual translation layers $k$ is set to 4. The maximum number of negative samples in two levels is both set to 512. The GNN model $f_e(\cdot ;\cal{G})$ is implemented by Node2vec or CompGCN. Please see Appendix~\ref{appendix_hyperparameters} for more details.

\noindent \textbf{Evaluation Protocol.} We evaluate the performance of our SMiLE on the link prediction task.
We regard the following two measurements as the evaluation metrics (~\citealp{SLiCE}; \citealp{DT-GCN}) of prediction performance: (1) Micro-F1 score; (2) AUC-ROC score.

\subsection{Main Results}

We compare our proposed SMiLE with various state-of-the-art models, and experimental results are summarized in Table~\ref{tab:comparision results}.
We reuse the results on FB15k-237 reported by \citet{SLiCE} for TransE, CompGCN and SLiCE. Clearly, we can observe that our proposed model SMiLE obtains competitive results compared with the baselines. Specifically, SMiLE performs better than relation-based method CompGCN which only models relational connection within a triple, emphasizing the contextual information learned from context schema is more effective in link prediction.
Furthermore, SMiLE outperforms the state-of-the-art baseline SLiCE(that shares the same backbone with SMiLE but is free of the schema context and global correlations between contexts) by a large margin on the FB15k, JF17k and HumanWiki datasets, but marginally lags behind on FB15k-237.

Compared to other datasets, the graph in FB15k-237 is much denser as the degree number of each entity is larger. In this case, models are more dependent on generalizable logic rules for KG inference. As SLiCE automatically learns meta-paths from contexts, it is quite helpful for link prediction. Besides, FB15k-237 dataset is reported to exist plenty of unpredictable links~\citep{missing_KGC}. Hence it is reasonable for the unsatisfactory result of SMiLE.

\subsection{Ablation Study}

We consider two ablated variants (contextual-level and global-level contrastive learning) of our model in the ablation study. The experimental results on FB15k and HumanWiki datasets are described in table~\ref{tab:Ablation Study}. We can observe that the full model(the third row) outperforms all those with a single component by a large margin on both micro-F1 and AUC-ROC scores, further certifying that semantic and structural information in contextual- and global-level both play fruitful contributions to SMiLE.

Moreover, we have an interesting observation that global-level information contributes more to the performance of the full model on FB15k than HumanWiki dataset. We believe that the graph of FB15k dataset is much denser, because each entity in it has a larger degree on average and naturally gains more information from its local neighbors. Under this circumstance, global-level information can be a significant promotion to KG embeddings.

\subsection{Impact of Negative Samples}

\begin{table}[!h]\small
\renewcommand{\arraystretch}{1.2}
 \centering
 \setlength{\tabcolsep}{2mm}
 \begin{tabular}{ccccc}\toprule
    \multirow{2}{*}{\textbf{Negatives}} & \multicolumn{2}{c}{\textbf{FB15k}} & \multicolumn{2}{c}{\textbf{HumanWiki}}
    \\\cmidrule(lr){2-3}\cmidrule(lr){4-5}
            & F1 & AUC-ROC  & F1 & AUC-ROC\\\midrule \specialrule{0em}{1.5pt}{1.5pt}\midrule
    Relation & 89.14 & 95.45 & 92.41 & 97.51 \\
    Schema & \textbf{90.76} & \textbf{96.53} & \textbf{93.40} & \textbf{97.92}
    \\\bottomrule
 \end{tabular}
 \caption{Performance of SMiLE with different kinds of negative samples on FB15k-237 and HumanWiki datasets.}
 \label{tab:Negative Samples}
\end{table}

In SMiLE, we adopt a contrastive learning strategy in the pre-train phase, which relies on the quality of negative samples. To verify whether our schema-guided sampling strategy obtains harder negatives, we compared it with a simpler relation-level sampling strategy, which randomly corrupts $h$ or $t$ in a positive triple $(s,r,o)$ with the constraint on entity type as follows:
\[
{\cal N}_s^{rel} = \big \{v\ |\ v \in {\cal E}, (s,*,v) \notin {\cal T}, t_v = t_o \big \},
\]
where $*$ means that there is no relation directly connecting entity $s$ and entity $v$.

As shown in Table~\ref{tab:Negative Samples}, by switching the negative sampling method from schema-level to relation-level, the micro-F1 score drops from 92.08\% to 90.77\%, and the AUC-ROC score drops from 97.23\% to 96.48\% on average, while it still leads to a competitive performance comparing to other KGE baselines. It is evident that the relation-level sampling strategy only focuses on an individual triple with type constraints on each entity, ignoring the context information of an entity. To summarize, the proposed schema-guided negative sampling strategy is capable of sampling effective hard negatives compared with traditional vanilla ones.

\subsection{Analysis on Discriminative Capacity} \label{Discriminative Capacity}

In this section, to further demonstrate the discriminative capacity of SMiLE on link prediction task, we visualize the distribution of positive and negative triple scores computed by SMiLE, and compare it with another GNN-based model Node2vec.

\begin{figure}[!t]
    \centering
    \subfloat[Node2vec on FB15k] {\includegraphics[width=1.5in] {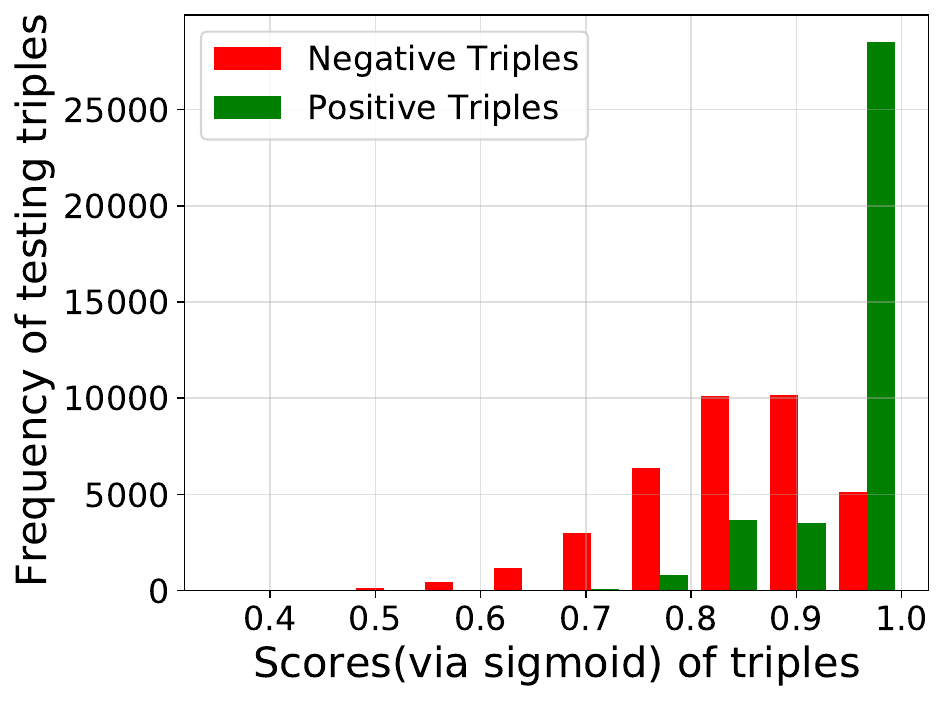} \label{Node2vec_FB15k_bar}}
    \subfloat[SMiLE on FB15k] {\includegraphics[width=1.5in] {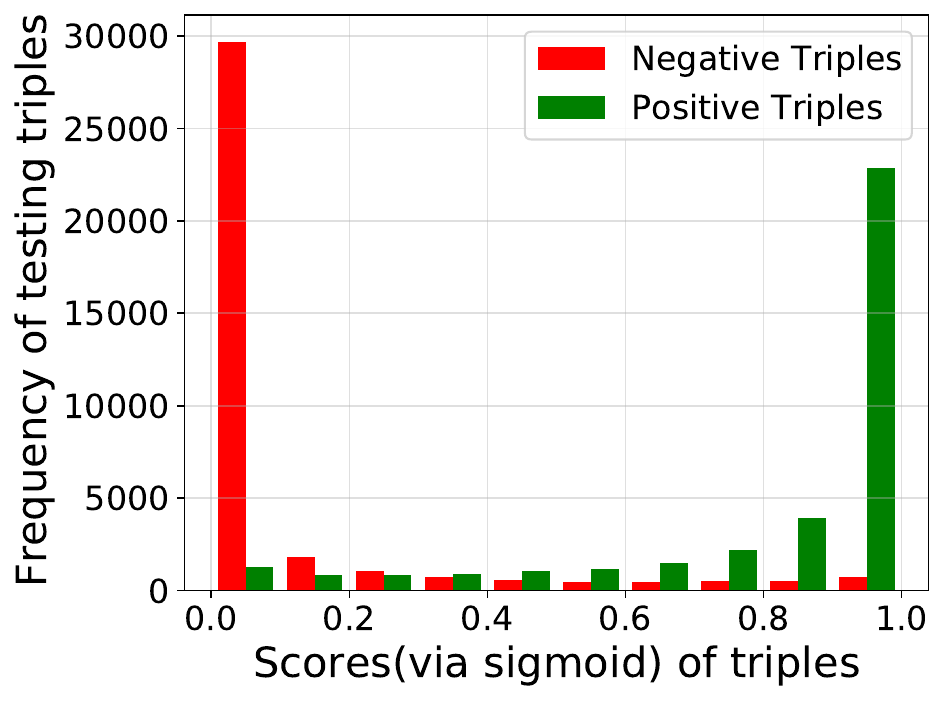} \label{SMiLE_FB15k_bar}} \\
    \subfloat[Node2vec on HumanWiki] {\includegraphics[width=1.5in] {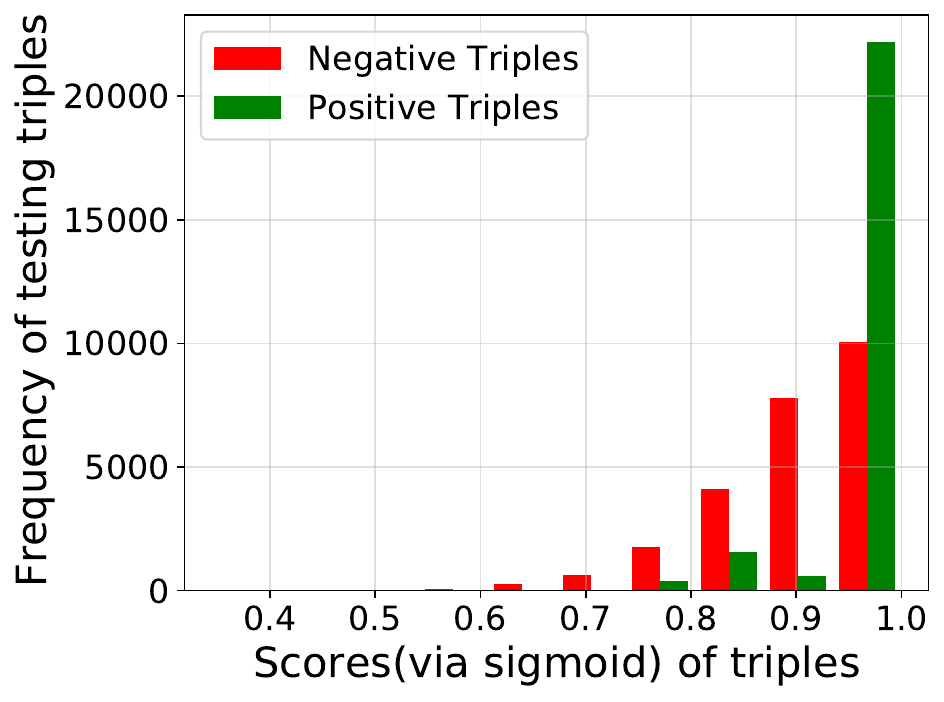} \label{Node2vec_HumanWiki_bar}}
    \subfloat[SMiLE on HumanWiki] {\includegraphics[width=1.5in] {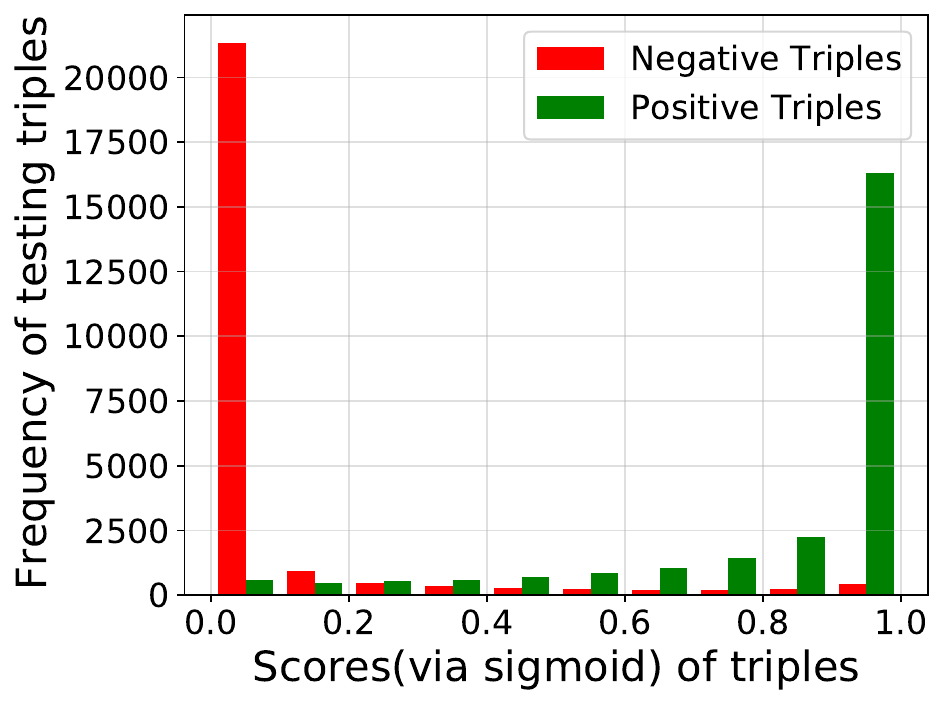} \label{SMiLE_HumanWiki_bar}}
    \caption{Histogram distribution of triple scores on FB15k and HumanWiki datasets.} 
    \label{fig:Distribution of Triple Scores}
\end{figure}

As shown in Figure~\ref{fig:Distribution of Triple Scores}, Node2vec model in \ref{Node2vec_FB15k_bar} and \ref{Node2vec_HumanWiki_bar} can not precisely discriminate positive triples and corrupted negative triples in test set, as a result of that a large number of negative triples still obtain high scores. Conversely, SMiLE in \ref{SMiLE_FB15k_bar} and \ref{SMiLE_HumanWiki_bar} increases the margin of distribution between positive triples and negative triples, offering strong evidence that our model brings positive triples closer while pushing negative triples farther away.

\subsection{Case Study}

\begin{table*}[!t]\small
\renewcommand{\arraystretch}{1.4} 
\centering
\setlength{\tabcolsep}{1.8mm}
\begin{tabular}{l|l|l|l}
\hline
 & \textbf{Entity Instance} & \textbf{Type Information} & \textbf{Context Schema}\\
\hline
\multirow{3}{*}{1} & \multirow{3}{*}{Warner Bros.} & business.employer & \textbf{Entity-typed Triples:}\\
& & award.award\_winner & \\
& & film.production\_company & (award\_winner, /place\_lived/location, administrative\_division) \\ \cline{1-3}
\multirow{2}{*}{2} & \multirow{2}{*}{California} & government.political\_district & \multirow{2}{*}{(award\_winner, /person/ethnicity, book\_subject)}\\
& & location.administrative\_division & \\ \cline{1-3}
\multirow{2}{*}{3} & \multirow{2}{*}{African Americans} & people.ethnicity & (administrative\_division, /location/containedby, book\_subject)\\
& & book.book\_subject & \\
\hline
\end{tabular}
\caption{Examples of representative entities on the test set of FB15k dataset with their detailed type information. Types of \textit{Warner Bros.}, \textit{California}, \textit{African Americans} and relations among them make up a context schema.}
\label{tab:Case Study}
\end{table*}

To examine the effectiveness and interpretability of our proposed model, we visualize the entity embeddings in 6 different contexts. We randomly select 6 tail entities from FB15k dataset, and for each tail entity we randomly sample 50 head entities that are connected to it via a relation. We visualize these entity embeddings computed with Node2vec and SMiLE, respectively.

\begin{figure}[!t]
    \centering
    \subfloat[Node2vec]{\includegraphics[width=1.5in]{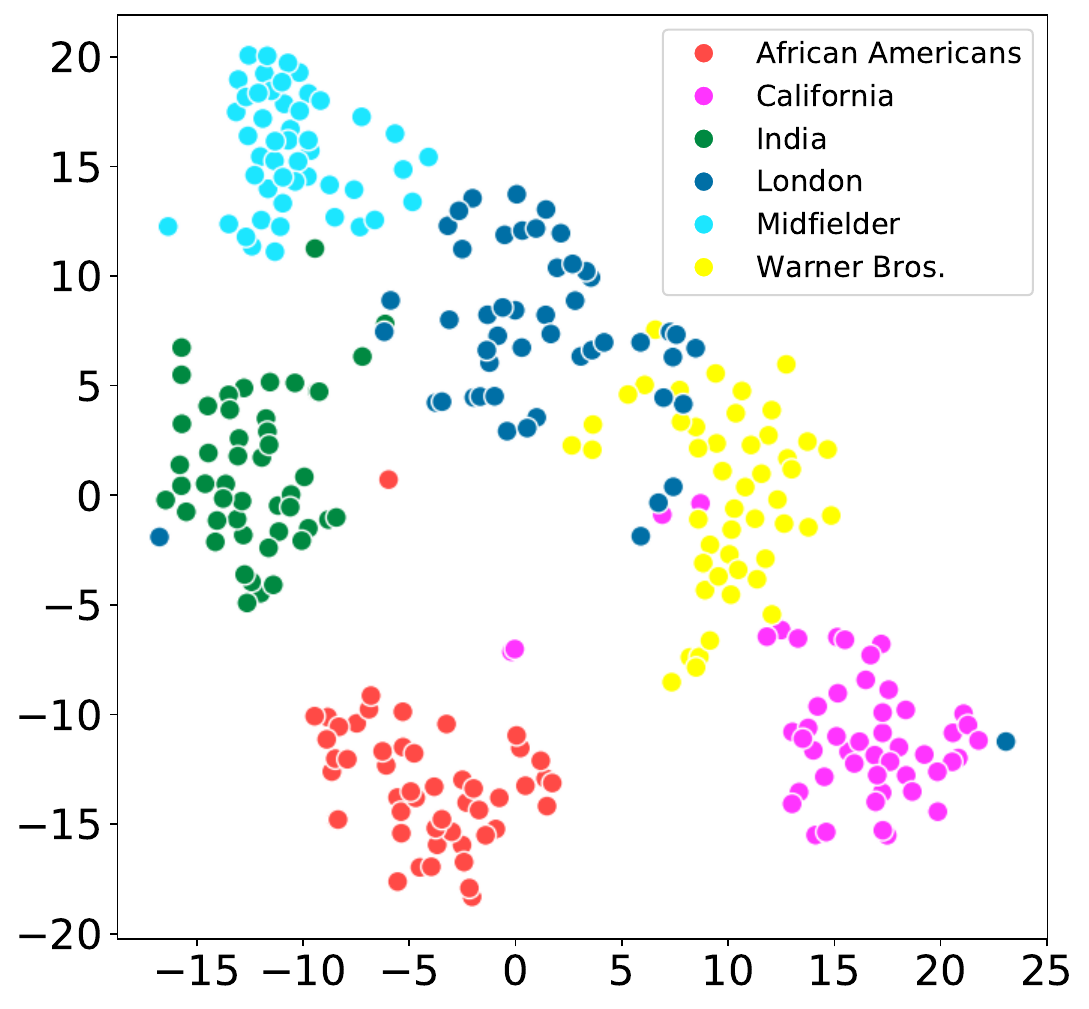} \label{Node2vec_vis}} 
    \subfloat[SMiLE]{\includegraphics[width=1.5in]{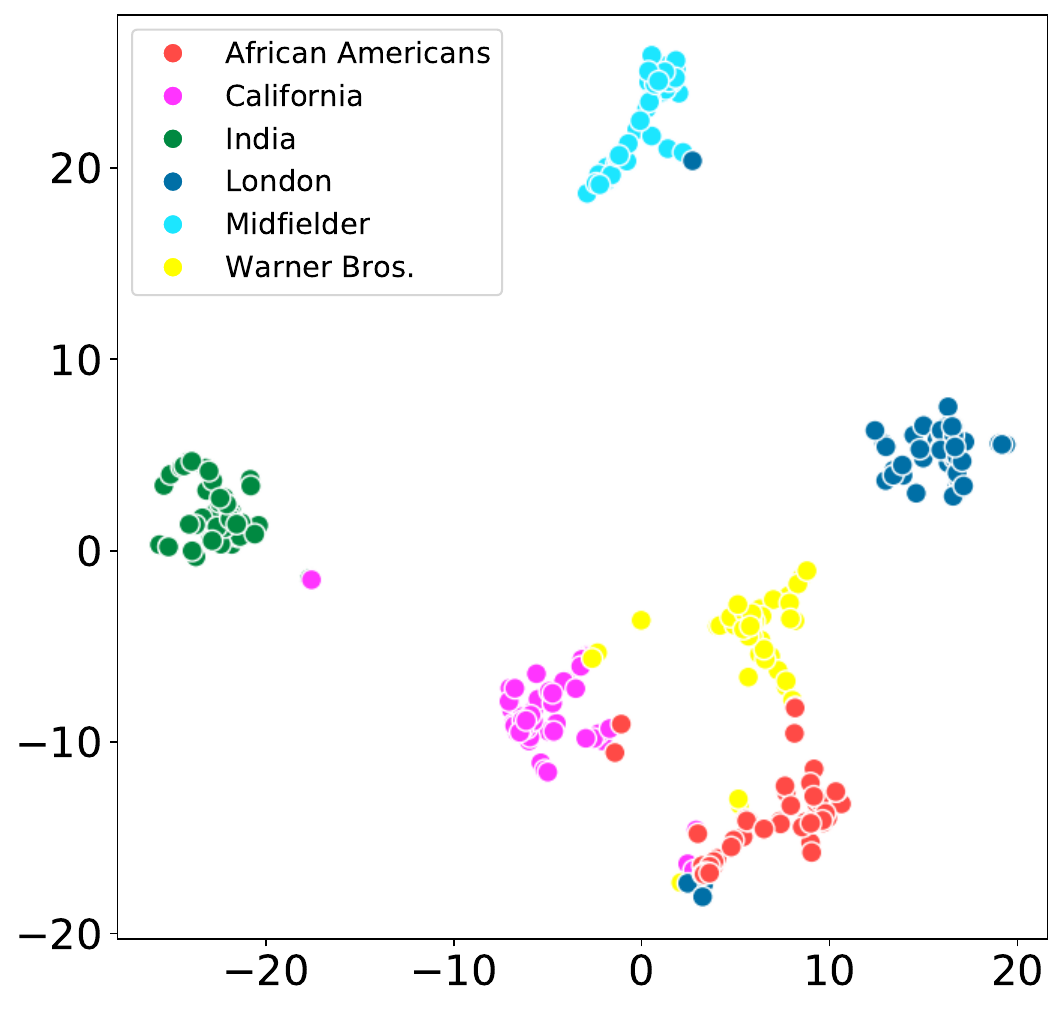} \label{SMiLE_vis}} 
    \caption{The visualization of entity embeddings on FB15k dataset using t-SNE\citep{t-SNE}. Points in same color indicate that they are head entities connected to the same tail entity via a relation.}
    \label{fig:embedding visulization}
\end{figure}

As shown in Figure~\ref{fig:embedding visulization}, model Node2vec in Figure~\ref{Node2vec_vis} can not separate entities in different contexts distinctly, especially there are some overlap between entities in context  \textit{Warner Bros.} and those in context \textit{London}. Conversely, entities in different contexts are well separated by utilizing SMiLE in Figure~\ref{SMiLE_vis} as an encoder. Moreover, the distance of entities within the same context is much closer, while the distribution of different contexts is much wider. Less overlap among clusters demonstrates that the proposed SMiLE effectively models the contextual information of entities while it distinguishes entities of different types more apart.

More concretely, we list more details of related type information in Table~\ref{tab:Case Study}. We can observe that entities of \textit{Warner Bros.}, \textit{California}, \textit{African Americans} and relations among them make up a bigger context schema. It is evident that there exist some semantic connections about "America" between them, hence distance among these clusters is closer.

\section{Conclusion}
In this paper, we propose SMiLE, a schema-augmented multi-level contrastive learning framework for knowledge graph link prediction. We identify the critical issue of conducting effective link prediction is how to model precise and consistent contextual information of entities in different contexts. We propose an approach to automatically extract the complete schema from a KG. To fully capture contextual information of entities, we first sample the two-level negatives and perform contrast estimation in contextual-level and global-level, and then fine-tune the representations of entities and relations to learn subtler knowledge. Empirical experiments on four benchmark datasets demonstrate that our proposed model effectively captures specific contextual information and correlations between different contexts of an entity. 

\section*{Limitations}
In this paper, we utilize KG schema as a prior constraint to capture contextual information. However, there are several limitations in our method: 1) The construction of schema relies on explicit type information of entities while some KGs lack them. A promising improvement is to model recapitulate type semantics by utilizing linguistic information of concepts and word embeddings to capture the similarity between entities. 2) The proposed negative sampling strategy may be time-consuming in large-scale KGs. For future work, a more effective way to incorporate schema contexts into both entity and relation representations is worth exploring.

\section*{Acknowledgements}
We would like to thank all the anonymous reviewers for their insightful and valuable comments. This work was supported by the National Key Research and Development Program of China (Grant No.2021ZD0113304), General Program of Natural Science Foundation of China (NSFC) (Grant No.62072346), Key R\&D Project of Hubei Province (Grant NO.2020BAA021, NO.2021BBA099, NO.2021BAA029) and Application Foundation Frontier Project of Wuhan (Grant NO.2020010601012168).

\label{sec:bibtex}

\bibliography{anthology,custom}

\begin{thebibliography}{37}
\expandafter\ifx\csname natexlab\endcsname\relax\def\natexlab#1{#1}\fi

\bibitem[{Ahrabian et~al.(2020)Ahrabian, Feizi, Salehi, Hamilton, and
  Bose}]{SANS}
Kian Ahrabian, Aarash Feizi, Yasmin Salehi, William~L. Hamilton, and
  Avishek~Joey Bose. 2020.
\newblock \href {https://doi.org/10.18653/v1/2020.emnlp-main.492} {Structure
  aware negative sampling in knowledge graphs}.
\newblock In \emph{Proceedings of the 2020 Conference on Empirical Methods in
  Natural Language Processing (EMNLP)}, pages 6093--6101, Online. Association
  for Computational Linguistics.

\bibitem[{Bordes et~al.(2013)Bordes, Usunier, Garc{\'{\i}}a{-}Dur{\'{a}}n,
  Weston, and Yakhnenko}]{TransE}
Antoine Bordes, Nicolas Usunier, Alberto Garc{\'{\i}}a{-}Dur{\'{a}}n, Jason
  Weston, and Oksana Yakhnenko. 2013.
\newblock Translating embeddings for modeling multi-relational data.
\newblock In \emph{Advances in Neural Information Processing Systems 26: 27th
  Annual Conference on Neural Information Processing Systems 2013. Proceedings
  of a meeting held December 5-8, 2013, Lake Tahoe, Nevada, United States},
  pages 2787--2795.

\bibitem[{Cao et~al.(2021)Cao, Ji, Lv, Li, Wen, and Zhang}]{missing_KGC}
Yixin Cao, Xiang Ji, Xin Lv, Juanzi Li, Yonggang Wen, and Hanwang Zhang. 2021.
\newblock \href {https://doi.org/10.18653/v1/2021.acl-long.534} {Are missing
  links predictable? an inferential benchmark for knowledge graph completion}.
\newblock In \emph{Proceedings of the 59th Annual Meeting of the Association
  for Computational Linguistics and the 11th International Joint Conference on
  Natural Language Processing (Volume 1: Long Papers)}, pages 6855--6865,
  Online. Association for Computational Linguistics.

\bibitem[{Chao et~al.(2021)Chao, He, Wang, and Chu}]{pairRE}
Linlin Chao, Jianshan He, Taifeng Wang, and Wei Chu. 2021.
\newblock \href {https://doi.org/10.18653/v1/2021.acl-long.336} {{P}air{RE}:
  Knowledge graph embeddings via paired relation vectors}.
\newblock In \emph{Proceedings of the 59th Annual Meeting of the Association
  for Computational Linguistics and the 11th International Joint Conference on
  Natural Language Processing (Volume 1: Long Papers)}, pages 4360--4369,
  Online. Association for Computational Linguistics.

\bibitem[{Diefenbach et~al.(2018)Diefenbach, Singh, and Maret}]{Diefenbach_QA}
Dennis Diefenbach, Kamal~Deep Singh, and Pierre Maret. 2018.
\newblock Wdaqua-core1: {A} question answering service for {RDF} knowledge
  bases.
\newblock In \emph{Companion of the The Web Conference 2018 on The Web
  Conference 2018, {WWW} 2018, Lyon , France, April 23-27, 2018}, pages
  1087--1091.

\bibitem[{Grover and Leskovec(2016)}]{Node2vec}
Aditya Grover and Jure Leskovec. 2016.
\newblock node2vec: Scalable feature learning for networks.
\newblock In \emph{Proceedings of the 22nd {ACM} {SIGKDD} International
  Conference on Knowledge Discovery and Data Mining, San Francisco, CA, USA,
  August 13-17, 2016}, pages 855--864. {ACM}.

\bibitem[{Hao et~al.(2019)Hao, Chen, Yu, Sun, and Wang}]{JOIE}
Junheng Hao, Muhao Chen, Wenchao Yu, Yizhou Sun, and Wei Wang. 2019.
\newblock Universal representation learning of knowledge bases by jointly
  embedding instances and ontological concepts.
\newblock In \emph{Proceedings of the 25th {ACM} {SIGKDD} International
  Conference on Knowledge Discovery {\&} Data Mining, {KDD} 2019, Anchorage,
  AK, USA, August 4-8, 2019}, pages 1709--1719.

\bibitem[{He et~al.(2020)He, Fan, Wu, Xie, and Girshick}]{MoCo}
Kaiming He, Haoqi Fan, Yuxin Wu, Saining Xie, and Ross~B. Girshick. 2020.
\newblock \href {https://doi.org/10.1109/CVPR42600.2020.00975} {Momentum
  contrast for unsupervised visual representation learning}.
\newblock In \emph{2020 {IEEE/CVF} Conference on Computer Vision and Pattern
  Recognition, {CVPR} 2020, Seattle, WA, USA, June 13-19, 2020}, pages
  9726--9735.

\bibitem[{Jain et~al.(2018)Jain, Kumar, {Mausam}, and
  Chakrabarti}]{TypeComplEx}
Prachi Jain, Pankaj Kumar, {Mausam}, and Soumen Chakrabarti. 2018.
\newblock \href {https://doi.org/10.18653/v1/P18-2013} {Type-sensitive
  knowledge base inference without explicit type supervision}.
\newblock In \emph{Proceedings of the 56th Annual Meeting of the Association
  for Computational Linguistics (Volume 2: Short Papers)}, pages 75--80,
  Melbourne, Australia. Association for Computational Linguistics.

\bibitem[{Jiang et~al.(2021{\natexlab{a}})Jiang, Jia, Fang, Shi, Lin, and
  Wang}]{PTHGNN}
Xunqiang Jiang, Tianrui Jia, Yuan Fang, Chuan Shi, Zhe Lin, and Hui Wang.
  2021{\natexlab{a}}.
\newblock Pre-training on large-scale heterogeneous graph.
\newblock In \emph{{KDD} '21: The 27th {ACM} {SIGKDD} Conference on Knowledge
  Discovery and Data Mining, Virtual Event, Singapore, August 14-18, 2021},
  pages 756--766. {ACM}.

\bibitem[{Jiang et~al.(2021{\natexlab{b}})Jiang, Lu, Fang, and Shi}]{CPT-KG}
Xunqiang Jiang, Yuanfu Lu, Yuan Fang, and Chuan Shi. 2021{\natexlab{b}}.
\newblock Contrastive pre-training of gnns on heterogeneous graphs.
\newblock In \emph{{CIKM} '21: The 30th {ACM} International Conference on
  Information and Knowledge Management, Virtual Event, Queensland, Australia,
  November 1 - 5, 2021}, pages 803--812.

\bibitem[{Lacroix et~al.(2018)Lacroix, Usunier, and Obozinski}]{ComplEx-N3}
Timoth{\'{e}}e Lacroix, Nicolas Usunier, and Guillaume Obozinski. 2018.
\newblock Canonical tensor decomposition for knowledge base completion.
\newblock In \emph{Proceedings of the 35th International Conference on Machine
  Learning, {ICML} 2018, Stockholmsm{\"{a}}ssan, Stockholm, Sweden, July 10-15,
  2018}, volume~80 of \emph{Proceedings of Machine Learning Research}, pages
  2869--2878. {PMLR}.

\bibitem[{Lao et~al.(2011)Lao, Mitchell, and Cohen}]{RWI_KB}
Ni~Lao, Tom Mitchell, and William~W. Cohen. 2011.
\newblock \href {https://aclanthology.org/D11-1049} {Random walk inference and
  learning in a large scale knowledge base}.
\newblock In \emph{Proceedings of the 2011 Conference on Empirical Methods in
  Natural Language Processing}, pages 529--539, Edinburgh, Scotland, UK.
  Association for Computational Linguistics.

\bibitem[{Le{-}Khac et~al.(2020)Le{-}Khac, Healy, and Smeaton}]{CRL2020}
Phuc~H. Le{-}Khac, Graham Healy, and Alan~F. Smeaton. 2020.
\newblock \href {https://doi.org/10.1109/ACCESS.2020.3031549} {Contrastive
  representation learning: {A} framework and review}.
\newblock \emph{{IEEE} Access}, 8:193907--193934.

\bibitem[{Lin et~al.(2018)Lin, Socher, and Xiong}]{Multi-Hop_Reasoning_RL}
Xi~Victoria Lin, Richard Socher, and Caiming Xiong. 2018.
\newblock \href {https://doi.org/10.18653/v1/D18-1362} {Multi-hop knowledge
  graph reasoning with reward shaping}.
\newblock In \emph{Proceedings of the 2018 Conference on Empirical Methods in
  Natural Language Processing}, pages 3243--3253, Brussels, Belgium.
  Association for Computational Linguistics.

\bibitem[{Lin et~al.(2015)Lin, Liu, Sun, Liu, and Zhu}]{TransR}
Yankai Lin, Zhiyuan Liu, Maosong Sun, Yang Liu, and Xuan Zhu. 2015.
\newblock Learning entity and relation embeddings for knowledge graph
  completion.
\newblock In \emph{Proceedings of the Twenty-Ninth {AAAI} Conference on
  Artificial Intelligence, January 25-30, 2015, Austin, Texas, {USA}}, pages
  2181--2187.

\bibitem[{Ma et~al.(2017)Ma, Ding, Jia, Wang, and Guo}]{TransT}
Shiheng Ma, Jianhui Ding, Weijia Jia, Kun Wang, and Minyi Guo. 2017.
\newblock Transt: Type-based multiple embedding representations for knowledge
  graph completion.
\newblock In \emph{Machine Learning and Knowledge Discovery in Databases -
  European Conference, {ECML} {PKDD} 2017, Skopje, Macedonia, September 18-22,
  2017, Proceedings, Part {I}}, volume 10534 of \emph{Lecture Notes in Computer
  Science}, pages 717--733.

\bibitem[{Meilicke et~al.(2019)Meilicke, Chekol, Ruffinelli, and
  Stuckenschmidt}]{BURL_KGC}
Christian Meilicke, Melisachew~Wudage Chekol, Daniel Ruffinelli, and Heiner
  Stuckenschmidt. 2019.
\newblock Anytime bottom-up rule learning for knowledge graph completion.
\newblock In \emph{Proceedings of the Twenty-Eighth International Joint
  Conference on Artificial Intelligence, {IJCAI} 2019, Macao, China, August
  10-16, 2019}, pages 3137--3143. ijcai.org.

\bibitem[{Nickel et~al.(2011)Nickel, Tresp, and Kriegel}]{RESCAL}
Maximilian Nickel, Volker Tresp, and Hans{-}Peter Kriegel. 2011.
\newblock A three-way model for collective learning on multi-relational data.
\newblock In \emph{Proceedings of the 28th International Conference on Machine
  Learning, {ICML} 2011, Bellevue, Washington, USA, June 28 - July 2, 2011},
  pages 809--816. Omnipress.

\bibitem[{Ouyang et~al.(2021)Ouyang, Huang, Chen, Tan, Liu, Sun, and Zhu}]{CCC}
Bo~Ouyang, Wenbing Huang, Runfa Chen, Zhixing Tan, Yang Liu, Maosong Sun, and
  Jihong Zhu. 2021.
\newblock \href {https://doi.org/10.18653/v1/2021.findings-emnlp.263}
  {Knowledge representation learning with contrastive completion coding}.
\newblock In \emph{Findings of the Association for Computational Linguistics:
  EMNLP 2021}, pages 3061--3073, Punta Cana, Dominican Republic. Association
  for Computational Linguistics.

\bibitem[{Richardson and Domingos(2006)}]{Markov_logic}
Matthew Richardson and Pedro~M. Domingos. 2006.
\newblock Markov logic networks.
\newblock \emph{Mach. Learn.}, 62(1-2):107--136.

\bibitem[{Rosso et~al.(2021)Rosso, Yang, Ostapuk, and
  Cudr{\'{e}}{-}Mauroux}]{RETA}
Paolo Rosso, Dingqi Yang, Natalia Ostapuk, and Philippe Cudr{\'{e}}{-}Mauroux.
  2021.
\newblock {RETA:} {A} schema-aware, end-to-end solution for instance completion
  in knowledge graphs.
\newblock In \emph{{WWW} '21: The Web Conference 2021, Virtual Event /
  Ljubljana, Slovenia, April 19-23, 2021}, pages 845--856.

\bibitem[{Schlichtkrull et~al.(2018)Schlichtkrull, Kipf, Bloem, van~den Berg,
  Titov, and Welling}]{R-GCN}
Michael~Sejr Schlichtkrull, Thomas~N. Kipf, Peter Bloem, Rianne van~den Berg,
  Ivan Titov, and Max Welling. 2018.
\newblock Modeling relational data with graph convolutional networks.
\newblock In \emph{The Semantic Web - 15th International Conference, {ESWC}
  2018, Heraklion, Crete, Greece, June 3-7, 2018, Proceedings}, volume 10843 of
  \emph{Lecture Notes in Computer Science}, pages 593--607.

\bibitem[{Shen et~al.(2021)Shen, Li, Wang, Li, and Zhang}]{DT-GCN}
Yuxin Shen, Zhao Li, Xin Wang, Jianxin Li, and Xiaowang Zhang. 2021.
\newblock Datatype-aware knowledge graph representation learning in hyperbolic
  space.
\newblock In \emph{{CIKM} '21: The 30th {ACM} International Conference on
  Information and Knowledge Management, Virtual Event, Queensland, Australia,
  November 1 - 5, 2021}, pages 1630--1639.

\bibitem[{Sun et~al.(2019)Sun, Deng, Nie, and Tang}]{RotatE}
Zhiqing Sun, Zhi{-}Hong Deng, Jian{-}Yun Nie, and Jian Tang. 2019.
\newblock Rotate: Knowledge graph embedding by relational rotation in complex
  space.
\newblock In \emph{7th International Conference on Learning Representations,
  {ICLR} 2019, New Orleans, LA, USA, May 6-9, 2019}.

\bibitem[{Toutanova and Chen(2015)}]{toutanova-chen-2015_FB15k-237}
Kristina Toutanova and Danqi Chen. 2015.
\newblock Observed versus latent features for knowledge base and text
  inference.
\newblock In \emph{Proceedings of the 3rd Workshop on Continuous Vector Space
  Models and their Compositionality}.

\bibitem[{Van~der Maaten and Hinton(2008)}]{t-SNE}
Laurens Van~der Maaten and Geoffrey Hinton. 2008.
\newblock Visualizing data using t-sne.
\newblock \emph{Journal of machine learning research}, 9(11).

\bibitem[{Vashishth et~al.(2020)Vashishth, Sanyal, Nitin, and
  Talukdar}]{CompGCN}
Shikhar Vashishth, Soumya Sanyal, Vikram Nitin, and Partha~P. Talukdar. 2020.
\newblock Composition-based multi-relational graph convolutional networks.
\newblock In \emph{8th International Conference on Learning Representations,
  {ICLR} 2020, Addis Ababa, Ethiopia, April 26-30, 2020}.

\bibitem[{Wang et~al.(2022)Wang, Zhao, Wei, and Liu}]{SimKGC}
Liang Wang, Wei Zhao, Zhuoyu Wei, and Jingming Liu. 2022.
\newblock \href {https://aclanthology.org/2022.acl-long.295} {{S}im{KGC}:
  Simple contrastive knowledge graph completion with pre-trained language
  models}.
\newblock In \emph{Proceedings of the 60th Annual Meeting of the Association
  for Computational Linguistics (Volume 1: Long Papers)}, pages 4281--4294,
  Dublin, Ireland. Association for Computational Linguistics.

\bibitem[{Wang et~al.(2021{\natexlab{a}})Wang, Agarwal, Ham, Choudhury, and
  Reddy}]{SLiCE}
Ping Wang, Khushbu Agarwal, Colby Ham, Sutanay Choudhury, and Chandan~K. Reddy.
  2021{\natexlab{a}}.
\newblock \href {https://doi.org/10.1145/3442381.3450060} {Self-supervised
  learning of contextual embeddings for link prediction in heterogeneous
  networks}.
\newblock In \emph{{WWW} '21: The Web Conference 2021, Virtual Event /
  Ljubljana, Slovenia, April 19-23, 2021}, pages 2946--2957. {ACM} / {IW3C2}.

\bibitem[{Wang et~al.(2021{\natexlab{b}})Wang, Huang, Wang, Yuan, Liu, He, and
  Chua}]{KGIN}
Xiang Wang, Tinglin Huang, Dingxian Wang, Yancheng Yuan, Zhenguang Liu,
  Xiangnan He, and Tat{-}Seng Chua. 2021{\natexlab{b}}.
\newblock Learning intents behind interactions with knowledge graph for
  recommendation.
\newblock In \emph{{WWW} '21: The Web Conference 2021, Virtual Event /
  Ljubljana, Slovenia, April 19-23, 2021}, pages 878--887.

\bibitem[{Wang et~al.(2021{\natexlab{c}})Wang, Liu, Han, and Shi}]{HeCo}
Xiao Wang, Nian Liu, Hui Han, and Chuan Shi. 2021{\natexlab{c}}.
\newblock Self-supervised heterogeneous graph neural network with
  co-contrastive learning.
\newblock In \emph{{KDD} '21: The 27th {ACM} {SIGKDD} Conference on Knowledge
  Discovery and Data Mining, Virtual Event, Singapore, August 14-18, 2021},
  pages 1726--1736. {ACM}.

\bibitem[{Wen et~al.(2016)Wen, Li, Mao, Chen, and Zhang}]{WenLMCZ16_JF17k}
Jianfeng Wen, Jianxin Li, Yongyi Mao, Shini Chen, and Richong Zhang. 2016.
\newblock On the representation and embedding of knowledge bases beyond binary
  relations.
\newblock In \emph{Proceedings of the Twenty-Fifth International Joint
  Conference on Artificial Intelligence, {IJCAI} 2016, New York, NY, USA, 9-15
  July 2016}, pages 1300--1307. {IJCAI/AAAI} Press.

\bibitem[{Xie et~al.(2016)Xie, Liu, and Sun}]{TKRL}
Ruobing Xie, Zhiyuan Liu, and Maosong Sun. 2016.
\newblock Representation learning of knowledge graphs with hierarchical types.
\newblock In \emph{Proceedings of the Twenty-Fifth International Joint
  Conference on Artificial Intelligence, {IJCAI} 2016, New York, NY, USA, 9-15
  July 2016}, pages 2965--2971.

\bibitem[{Xiong et~al.(2017)Xiong, Hoang, and Wang}]{DeepPath}
Wenhan Xiong, Thien Hoang, and William~Yang Wang. 2017.
\newblock \href {https://doi.org/10.18653/v1/D17-1060} {{D}eep{P}ath: A
  reinforcement learning method for knowledge graph reasoning}.
\newblock In \emph{Proceedings of the 2017 Conference on Empirical Methods in
  Natural Language Processing}, pages 564--573, Copenhagen, Denmark.
  Association for Computational Linguistics.

\bibitem[{Yang et~al.(2015)Yang, Yih, He, Gao, and Deng}]{DistMult}
Bishan Yang, Wen{-}tau Yih, Xiaodong He, Jianfeng Gao, and Li~Deng. 2015.
\newblock Embedding entities and relations for learning and inference in
  knowledge bases.
\newblock In \emph{3rd International Conference on Learning Representations,
  {ICLR} 2015, San Diego, CA, USA, May 7-9, 2015, Conference Track
  Proceedings}.

\bibitem[{Zhang et~al.(2019)Zhang, Tay, Yao, and Liu}]{QuatE}
Shuai Zhang, Yi~Tay, Lina Yao, and Qi~Liu. 2019.
\newblock Quaternion knowledge graph embeddings.
\newblock In \emph{Advances in Neural Information Processing Systems 32: Annual
  Conference on Neural Information Processing Systems 2019, NeurIPS 2019,
  December 8-14, 2019, Vancouver, BC, Canada}, pages 2731--2741.

\end{thebibliography}
\bibliographystyle{acl_natbib}
\clearpage
\appendix

\section{More details about Experiment} \label{sec:More details about Experiment}
\subsection{Datasets}
In this paper, we utilize FB15k~\citep{TransE}, FB15k-237~\citep{toutanova-chen-2015_FB15k-237}, JF17k~\citep{TransE} and HumanWiki~\citep{RETA} to conduct experiments. The FB15k-237 dataset can be download on \url{https://github.com/malllabiisc/CompGCN/tree/master/data_compressed} and the FB15k, JF17k and HumanWiki datasets with their corresponding type information sets are taken from RETA~\citep{RETA} and can be download on \url{http://bit.ly/3t2WFTE}.

\subsection{Baselines}
The results of all baselines are obtained with their original implementations. For all baselines, we set the embedding dimension to 128. The other parameters that are not mentioned follow their original official settings.

We use the implementations of TransE and ComplEx-N3 provided in KGEmb\footnote{\url{https://github.com/HazyResearch/KGEmb}} and we adopt the origin implementation of TransR provided in OpenKE\footnote{\url{https://github.com/thunlp/OpenKE}}. We use the original settings of TypeComplEx\footnote{\url{https://github.com/dair-iitd/KBI/tree/master/kbi-pytorch}} and PairRE\footnote{\url{https://github.com/DeepGraphLearning/KnowledgeGraphEmbedding}} for the evaluation. The implementation of SANS\footnote{\url{https://github.com/kahrabian/SANS}} is based on DistMult with self-adversarial approach. CompGCN\footnote{\url{https://github.com/malllabiisc/CompGCN}} is implemented by using the DistMult score function with multiplication composition operator. For Node2vec\footnote{\url{https://github.com/aditya-grover/node2vec}}, we sample 10 random walks with the walk length of 80. In SLiCE\footnote{\url{https://github.com/pnnl/SLiCE}}, we use global embedding feature from Node2vec and both the number of self-attention heads and contextual translation layers are set to 4.

\subsection{Hyperparameters} \label{appendix_hyperparameters}

\begin{table}[!h]\small
\renewcommand{\arraystretch}{1.2}
\centering
\setlength{\tabcolsep}{1.4mm}
\begin{tabular}{l|c}
\toprule
\textbf{Hyperparameters} & \textbf{Values}\\
\midrule
Temperature $\tau$ & \{0.5, 0.6, 0.7, 0.8, 0.9\} \\
Learning rate & \{0.0001, 0.001, 0.03, 0.1\} \\
Balancing coefficient $\lambda$ & \{0.2, 0.4, 0.6, 0.8\} \\
Queued Negative Batches $n$ & \{1, 2, 5, 8, 10\} \\
Context subgraph size $|{\cal{E}}_c|$ & \{6, 12, 18\} \\
Embedding dimension & \{64, 128, 256, 512\} \\
Batch size(Pre-train) & \{128, 256, 512, 1024, 2048\} \\
Batch size(Fine-tune) & \{64, 128, 256, 512\} \\
epoch(Pre-train) & \{10, 15, 20\} \\
epoch(Fine-tune) & \{10, 15, 20\} \\
\bottomrule
\end{tabular}
\caption{Details of hyperparameters.}
\label{tab:hyperparameters}
\end{table}

To select the best hyperparameters for our model, we conduct the grid search on hyperparameters listed in Table~\ref{tab:hyperparameters} using the validation data. We set the embedding dimension on all datasets to 128. For batch size of pre-training, we use 1024, 2048, 512 and 1024 for FB15k, FB15k-237, JF17k and HumanWiki datasets. For fine-tuning batch size, we all set to 256. For schema frequency threshold, we use 700, 1000, 70 and 50 for FB15k, FB15k-237, JF17k and HumanWiki datasets. The queued negatives hyperparameter $n$ is set to 2 for all datasets. The maximum number of entities in a context subgraph is set to 12 on FB15k-237 and 6 on the other three datasets. We adopt multiplication as the implementation of entity-relation composition operation $\phi(\cdot)$.

Each epoch in pre-train phase takes $\sim$ 70s for FB15k, $\sim$ 130s for FB15k-237, $\sim$ 54s for JF17k and $\sim$ 600s for HumanWiki dataset.

\section{Additional Analysis Results} \label{sec:More Analysis Results}

\subsection{Analysis on Hyper-parameters}

\begin{figure}[!h]
    \centering
    \subfloat[Micro-F1]{\includegraphics[width=1.5in]{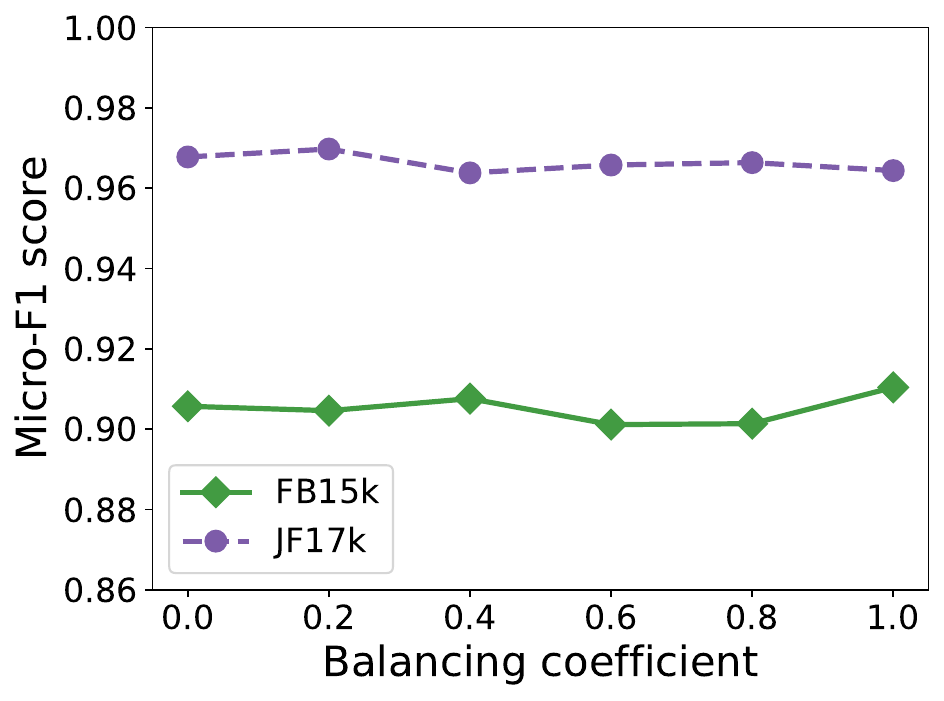} \label{balancing_coefficient_F1}} 
    \subfloat[AUC-ROC]{\includegraphics[width=1.5in]{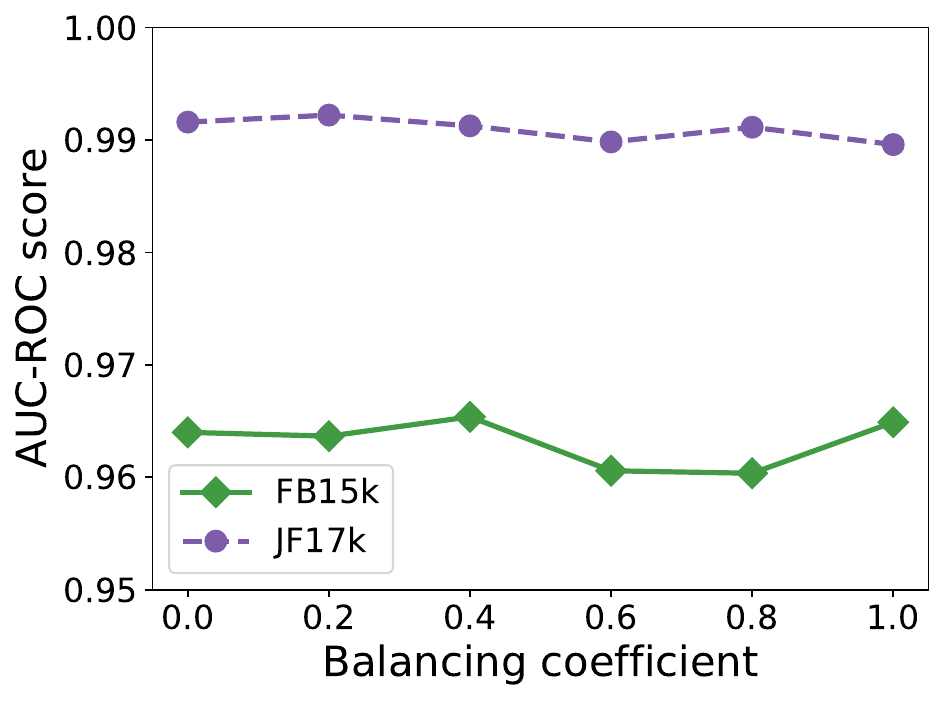} \label{balancing_coefficient_AUCROC}} 
    \caption{Effects of balancing coefficient $\lambda$ on datasets FB15k and JF17k with SMiLE.}
    \label{fig:balancing coefficient}
\end{figure}

\begin{table}[!h]\small
\renewcommand{\arraystretch}{1.2}
 \centering
 \setlength{\tabcolsep}{1.6mm}
 \begin{tabular}{ccccc}\toprule
    \textbf{Frequency} & \textbf{Schema} & \textbf{Coverage} & \multicolumn{2}{c}{\textbf{FB15k}}
    \\\cmidrule(lr){4-5}
           \textbf{Threshold $\alpha$} & \textbf{Size} & \textbf{Ratio} & F1 & AUC-ROC\\\midrule \specialrule{0em}{1.5pt}{1.5pt}\midrule
    400 & 10469 & 0.044\% & 87.97 & 94.52 \\
    600 & 5532 & 0.023\% & 89.22 & 95.47 \\
    700 & 4239 & 0.018\% & \textbf{90.76} & \textbf{96.54} \\
    800 & 3528 & 0.015\% & 89.04 & 95.47
    \\\bottomrule
 \end{tabular}
 \caption{Performance of SMiLE trained with schema in different scale on FB15k. Coverage Ratio indicates the ratio of filtered entity-typed triples to all candidate ones.}
 \label{tab:Different Schema Size}
\end{table}

In Figure~\ref{fig:balancing coefficient}, we display how the balancing coefficient $\lambda$ affects the performance of SMiLE on FB15k and JF17k datasets.

As mentioned in Section~\ref{Schema Construction}, we adopt the frequency threshold $\alpha$ to filter those meaningless entity-typed triples. In Table~\ref{tab:Different Schema Size}, we report the performance of SMiLE trained with different schema in different scales on FB15k dataset.

\subsection{Ablation on queued pre-batch negatives}

\begin{table}[!h]\small
\renewcommand{\arraystretch}{1.2}
 \centering
 \setlength{\tabcolsep}{1.6mm}
 \begin{tabular}{ccccc}\toprule
    \multirow{2}{*}{\textbf{Model}} & \multicolumn{2}{c}{\textbf{FB15k}} & \multicolumn{2}{c}{\textbf{HumanWiki}}
    \\\cmidrule(lr){2-3}\cmidrule(lr){4-5}
            & F1 & AUC-ROC  & F1 & AUC-ROC\\\midrule \specialrule{0em}{1.5pt}{1.5pt}\midrule
    w/o pre-batch & 90.58 & 96.42 & 92.77 & 97.62 \\
    w/ pre-batch & \textbf{90.76} & \textbf{96.53} & \textbf{93.40} & \textbf{97.92}
    \\\bottomrule
 \end{tabular}
 \caption{Performance of SMiLE in "without queued pre-batch negatives" and "full" modes on FB15k and HumanWiki datasets.}
 \label{tab:Without dynamic queue}
\end{table}

To explore how much the dynamic queue for the negatives contribute, we report the experimental results in Table~\ref{tab:Without dynamic queue}. We can observe that combining pre-batch negatives in contrastive learning is a promotion to the model performance.

\subsection{Effect of pre-training}

\begin{table}[!h]\small
\renewcommand{\arraystretch}{1.2}
 \centering
 \setlength{\tabcolsep}{1.6mm}
 \begin{tabular}{ccccc}\toprule
    \multirow{2}{*}{\textbf{Model}} & \multicolumn{2}{c}{\textbf{FB15k}} & \multicolumn{2}{c}{\textbf{HumanWiki}}
    \\\cmidrule(lr){2-3}\cmidrule(lr){4-5}
            & F1 & AUC-ROC  & F1 & AUC-ROC\\\midrule \specialrule{0em}{1.5pt}{1.5pt}\midrule
    SMiLE$_{w/o PT}$ & 65.84 & 60.46 & 65.22 & 59.85 \\
    SMiLE & \textbf{90.76} & \textbf{96.53} & \textbf{93.40} & \textbf{97.92}
    \\\bottomrule
 \end{tabular}
 \caption{Performance of SMiLE in "without pre-train" and "full" modes on FB15k and HumanWiki datasets.}
 \label{tab:Without pre-train}
\end{table}

To demonstrate the effect of pre-training phase on capturing the contextual knowledge of entities, we disable the pre-train phase from SMiLE(denoted as SMiLE$_{w/o PT}$) and only conduct fine-tune phase for link prediction. We report the result in Table~\ref{tab:Without pre-train}, and we can observe that without pre-trained entity embeddings with contextual knowledge, the performance of SMiLE decreased on both FB15k and HumanWiki datasets.

\subsection{Additional Results of Discriminative Capacity}

\begin{figure}[!h]
    \centering
    \subfloat[CompGCN on FB15k] {\includegraphics[width=1.5in] {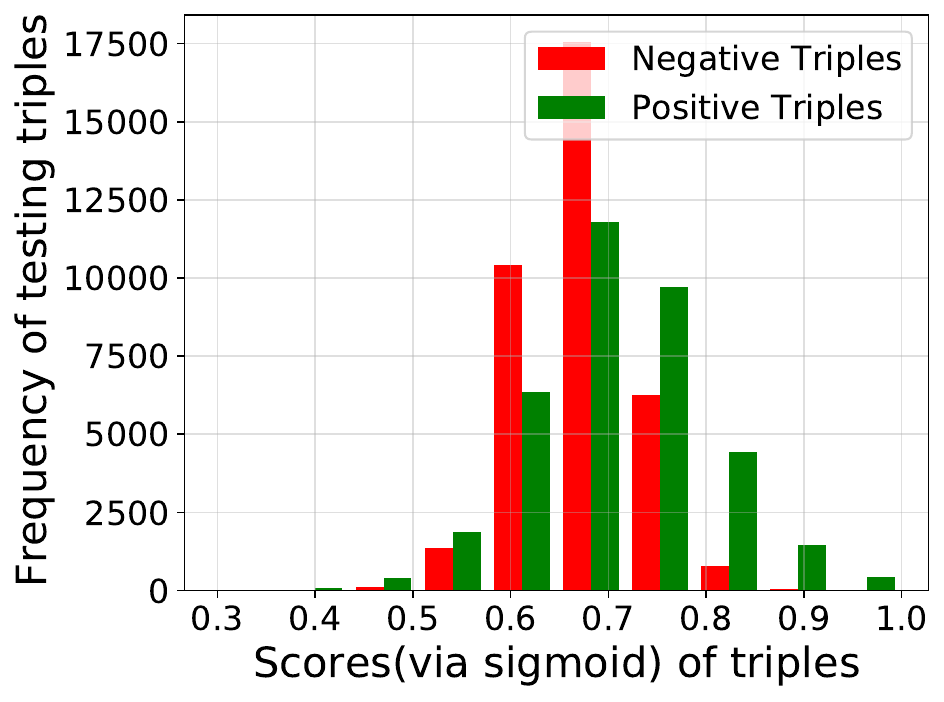} \label{CompGCN_FB15k_bar}}
    \subfloat[SMiLE on FB15k] {\includegraphics[width=1.5in] {smile_bar_fb15k.pdf} \label{SMiLE_FB15k_bar_apdx}} \\
    \subfloat[CompGCN on HumanWiki] {\includegraphics[width=1.5in] {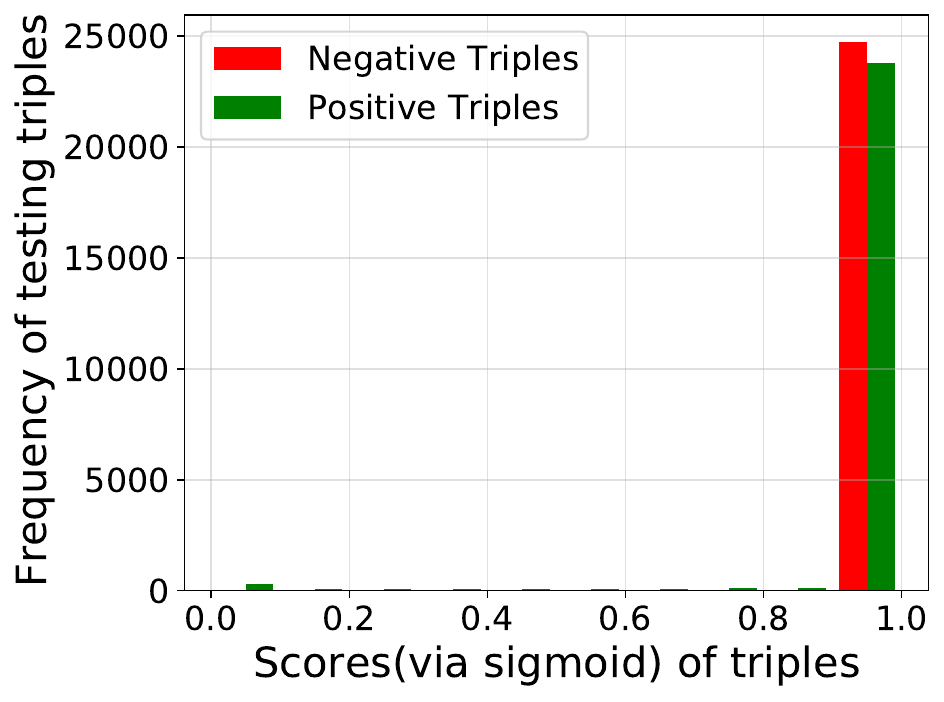} \label{CompGCN_HumanWiki_bar}}
    \subfloat[SMiLE on HumanWiki] {\includegraphics[width=1.5in] {smile_bar_humanwiki.pdf} \label{SMiLE_HumanWiki_bar_apdx}}
    \caption{Histogram distribution of triple scores computed with CompGCN and SMiLE respectively.}
    \label{fig:Distribution of Triple Scores on CompGCN}
\end{figure}

To supplement the analysis in \ref{Discriminative Capacity} and further demonstrate the discriminative capacity of our proposed SMiLE on link prediction, in Figure~\ref{fig:Distribution of Triple Scores on CompGCN} we visualize the distribution of triple scores computed with state-of-the-art GNN-based multi-relational model CompGCN~\citep{CompGCN}.

\end{document}